%% file: main.tex
\pdfoutput=1

\documentclass[11pt]{article}

\usepackage[]{acl}

\usepackage{times}
\usepackage{latexsym}
\usepackage{xcolor}
\usepackage{graphicx}
\usepackage{soul}
\usepackage{url}
\usepackage{booktabs}
\usepackage{amsmath}
\usepackage{cuted}
\usepackage{hyperref}

\usepackage[T1]{fontenc}

\usepackage[utf8]{inputenc}

\usepackage{microtype}

\title{XDLM: Cross-lingual Diffusion Language Model for Machine Translation}

\author{Linyao Chen$^1$, Aosong Feng$^2$, Boming Yang$^1$, Zihui Li$^1$ \\
$^1$University of Tokyo, $^2$Yale University\\
{chen-linyao217@g.ecc.u-tokyo.ac.jp}
}

\begin{document}
\maketitle
\begin{abstract}
Recently, diffusion models have excelled in image generation tasks and have also been applied to neural language processing (NLP) for controllable text generation. However, the application of diffusion models in a cross-lingual setting is less unexplored. Additionally, while pretraining with diffusion models has been studied within a single language, the potential of cross-lingual pretraining remains understudied. To address these gaps, we propose XDLM, a novel Cross-lingual diffusion model for machine translation, consisting of pretraining and fine-tuning stages.
In the pretraining stage, we propose TLDM, a new training objective for mastering the mapping between different languages; in the fine-tuning stage, we build up the translation system based on the pretrained model. We evaluate the result on several machine translation benchmarks and outperformed both diffusion and Transformer baselines. Our code is available in \href{https://github.com/Amayama/XDLM}{https://github.com/Amayama/XDLM}.

\end{abstract}

\input{1_introduction}

\input{methodology}

\input{experiment}

\input{ablation}

\section{Conclusion and Future Work}
In this study, we propose an innovative architecture that integrates cross-lingual pretraining into diffusion-based text generation. This is achieved through a carefully designed pretraining task. We compare our model with some previous works under automated evaluation method. 
Looking forward, we plan to extend our model to include additional languages, with the aim of constructing a robust multilingual model capable of handling more extensive cross-lingual translation tasks.
\bibliography{anthology,custom}
\bibliographystyle{acl_natbib}

\appendix



\end{document}

%% file: 1_introduction.tex
\section{Introduction}
\label{sec:introduction}

Diffusion-based generative models, or diffusion models \cite{ho2020denoising}, have recently demonstrated substantial potential for generating high-quality output in computer vision (CV). Furthermore, several recent studies have explored their application in natural language processing (NLP), including generation tasks such as machine translation, text summarization, and controllable text generation \cite{li2022diffusion, zheng2023reparameterized, gao2022difformer}. Notably, GENIE \cite{lin2022genie} proposes the use of pretraining on diffusion models, leveraging large English corpora and subsequent fine-tuning on downstream tasks. However, there is a lack of research investigating the cross-lingual application of diffusion models, particularly in the context of pretraining. 

There are two types of diffusion models: discrete and continuous. Some works focused on the discrete nature of text and have attempted to extend diffusion models to generate high-quality text. The discrete diffusion \cite{austin2021structured,hoogeboom2021argmax} model was initially proposed to generate text samples, solved by denoising and resetting the mask state for each token step by step. In the other hand, the continuous diffusion model \cite{li2022diffusion} was introduced later, which added additional embedding and rounding steps to transform discrete tokens into continuous latent representations, enabling gradient-based methods for controllable text generation. Then, GENIE model \cite{lin2022genie} involves integrating the diffusion model with a Transformer-based model, which resulted in a large-scale language pre-training model based on the diffusion framework.  Furthermore, the Difformer model \cite{gao2022difformer} has improved the existing diffusion methods by updating the loss function, adding a layer normalization and a noise factor, to establish a more stable diffusion process. \cite{zheng2023reparameterized} introduce a trick of reparameterization to the discrete diffusion, contributing to a simplified training process and a flexible sampling process. Inspired by GENIE, we propose to apply pretraining in a cross-lingual setting with continuous diffusion. 


In this paper, we examine the properties of a large-scale multilingual corpus and propose the implementation of cross-lingual pre-training denoising tasks to construct a framework for a cross-lingual diffusion model, termed as Cross-Lingual Diffusion Language Model (XDLM). XDLM specifically designs a cross-lingual pre-training task and corresponding objective for  multilingual data, enabling the diffusion model to comprehend the mapping relationships between various languages. To the best of our knowledge, this is the inaugural attempt to introduce the concept of cross-lingual pretraining to diffusion-based models. 


The principal contributions of this work can be summarized as follows:
\begin{itemize}
\item We introduce XDLM, the first architecture to the best of our knowledge aiming to incorporate cross-lingual pretraining into diffusion-based text generation.
\item We propose a pre-training task, Translation Diffusion Language Modeling (TDLM), along with a corresponding loss function. These enhancements augment the model's capacity to capture contextual correlations across various language domains. We also provide a discussion on potential issues. 
\end{itemize}

%% file: methodology.tex
\section{Cross-Lingual Diffusion Language Model}
\label{sec:XDLMusion}


In this section, we present the Cross-lingual Diffusion Language Model (XDLM), which incorporates a pretraining phase on cross-lingual data, utilizing diffusion techniques for the purpose of non-autoregressive machine translation, and a fine-tuning phase generating corresponding text from one language to another language based on the pretrained model.





\textbf{Non-AutoRegressive (NAR) Machine Translation}
In machine translation, given the input sequence from a source language $X=\{x_1,x_2,…,x_{|X|}\}$, the task is to generate the output sequence of the translation in the target language $Y=\{y_1, y_2,…, y_{|Y|}\}$. In this work, we focus on the Non-AutoRegressive (NAR) translation setting with the diffusion model. Typically, it has the following conditional probability:  
$$
p_{\theta}(Y|X)=\prod_{i=1}^{|Y|} p_{\theta}(y_i|X).
$$

Unlike AutoRegressive (AR) text generation, all tokens $y_i$$(0\leq i \leq |Y|)$ in the generated sequence $Y$ are predicted concurrently. The generation solely depends on the input sequence $X$, without any dependency on preceding tokens. This attribute presents a challenge in determining the length of the generated sequence. To address this issue, the length prediction of the output sequence is introduced as an auxiliary task \cite{gu2017non}. And the training loss is defined as a weighted sum between the translation loss and the length prediction loss. We apply length prediction in our finetuning phase following RDM(\citet{zheng2023reparameterized}).

\textbf{Diffusion Models}
The Denoising Diffusion Probabilistic Model (DDPM) \cite{ho2020denoising} is a parametrized Markov chain, and it is trained using variational inference to generate samples that match the original input data. 
The diffusion process comprises a noise-adding forward process and a noise-removing backward process, both of which can be viewed as discrete-time Markov processes. During the forward process, the model gradually introduces random noise with different scheduled variance $\beta_1,...,\beta_t$, with the aim of generating a standard Gaussian noise $x_t$ after $t$ turns. This can be formalized as follows:
$$
q(x_{t+1}|x_t)=\mathcal{N}(x_{t+1};\sqrt{1-\beta_{t+1}}x_t,\beta_{t}\mathbf{I}).
$$

The backward process, the reverse of the forward process, attempts to reconstruct the target sequence from the standard noise. Like the forward process, this procedure is also applied incrementally and can be formalized as follows:

$$
    p(x_{t-1}|x_t)=\mathcal{N}(x_{t-1};\mu_{\theta}^{t-1},\sigma_{\theta}^{t-1}),
$$
$$
    \mu_{\theta}^{t-1}=\frac{1}{\sqrt{\alpha_{t}}}(x_t-\frac{\beta_{t}}{\sqrt{1-\overline(\alpha_{t})}}z_{\theta}(x_{t},t)), 
$$
$$
    \sigma_{\theta}^{{t-1}}=\sqrt{\frac{1-\overline{\alpha_{t-1}}}{1-\overline{\alpha_{t}}}\dot \beta_{t}}.
$$

where $\alpha_t=1-\beta_t, \overline{\alpha_{t}}=\prod_{i=1}^t \alpha_{i}$ and $z_\theta$ comes from the prediction of model parameterized by $\theta$. 
In this work, we apply discrete diffusion for text generating and cross-lingual translation. Based on \citet{zheng2023reparameterized}, we follow the proposed discrete diffusion model with the following routing mechanism.

$$
    x_{t-1}, v_{t-1} \sim q(x_{t-1},v_{t-1}|x_t,x_0) \\
$$
$$
    q(v_{t-1}|x_t,x_0)=q(v_{t-1})=Bernoulli(\lambda) \\
$$
\begin{align*}
    & q(x_{t-1}|v_{t-1},x_t,x_0) = \nonumber \\ 
    &\qquad\left\{
    \begin{aligned}
        &v_{t-1}x_t+(1-v^{(1)}_{t-1})q_{\text{noise}}, \quad &&\text{if } x_t = x_0 \\
        &v_{t-1}x_0+(1-v_{t-1}^{(2)})q_{\text{noise}}(x_t), \quad &&\text{if } x_t \neq x_0
    \end{aligned}
    \right.
\end{align*}

Which models the joint distribution over both $x$ and $v$, where  $q_{noise}(x_t)=\beta_{t}x_t+(1-\beta_{t})q_{noise}$ and . The sampling process here also takes the reparameterized method, which improves flexibility and expressiveness compared to the vanilla multinomial diffusion process.

\begin{figure*}[htbp]
\centering
\includegraphics[width=\textwidth]{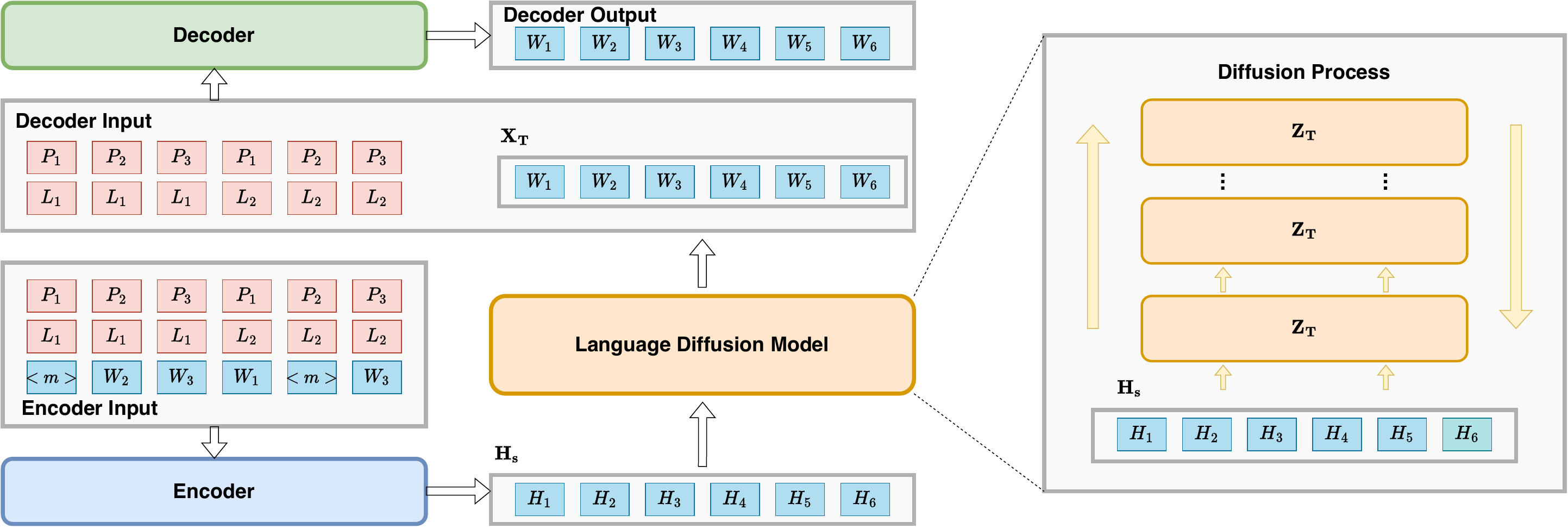}
\caption{The workflow of Translation Diffusion Language Modeling.}
\label{fig:my_label}
\end{figure*}
\textbf{Translation Diffusion Language Modeling (TDLM)}
Unlike previous diffusion model objectives for language modeling that primarily concentrate on monolingual data, we target to exploit cross-lingual modeling capabilities from parallel datasets. Consequently, we propose a pretraining process named Translation Diffusion Language Modeling (TDLM), aiming at enhancing cross-lingual pretraining with diffusion models. As illustrated in Figure 1, we first concatenate both source and target sentences and generate the corresponding language and position embedding sequences, and then stack them as the input to the encoder and diffusion model. 
The language and position embedding sequence is also introduced to the decoder, helping the model to map latent vectors to the generated sentences.
In a similar vein to \citet{lin2023text}, we random mask 15\% of the tokens to the input as \cite{lample2019cross} designed, tasking the model with predicting the noise and its surrounding text based on the cross-lingual context. This denoising setting assists the model in grasping the cross-lingual context.

\textbf{Translation Model Setting}
In our translation task, we employ a pretrained model based on Translation Diffusion Language Modeling (TDLM) during the fine-tuning stage. This model serves as a robust foundation for comprehending cross-lingual mapping relationships. The Diffusion model operates within an encoder-decoder architecture, where we utilize sentences from both the source and target domains to construct encoder inputs and target inputs, along with their corresponding position embeddings. Furthermore, the language embeddings of the input and output align with the languages presented in the source and target domains. Tokens from the same language as the pretraining process share identical embeddings, facilitating the model's rapid acclimatization to the source and target domains of the translation task.

%% file: experiment.tex
\section{Experiments}
\label{sec:experiments}

\subsection{Baselines and Datasets}
We conduct a large cross-lingual corpus and two standard benchmarks for cross-lingual translation, which are introduced as follows, and the information of each dataset is shown in Table \ref{table:stat}.
\begin{itemize}
\item Opus-ENDE\footnote{\url{https://opus.nlpl.eu/}}: This dataset comprises a large volume of English-German sentence pairs.
\item IWSLT14 DE-EN \cite{cettolo-etal-2014-report}: This benchmark is specifically employed for the German to English translation task.
\item WMT14 EN-DE \cite{bojar-etal-2014-findings}: This benchmark is designated for the English to German translation task.
\end{itemize}

\begin{table}[htbp]
\centering
\footnotesize
\begin{tabular*}{0.5\textwidth}{lcccc}
\toprule
 \textbf{Dataset} & \textbf{Usage} & \textbf{Train} & \textbf{Test} & \textbf{Valid}\\
\midrule
Opus-ENDE & pretrain & 9,323,066 & 5,000 & 5,000 \\
IWSLT14 & finetune & 160,240 & 6,750 & 7,283 \\
WMT14 & finetune & 4,496,988 & 3,003 & 3,000 \\
\bottomrule
\end{tabular*}
\caption{Dataset statistics.}
\label{table:stat}
\end{table}

We use the origin datasets with the same split and do not apply distillation on the dataset. Besides, we follow the data processing introduced by fairseq \footnote{\url{https://github.com/facebookresearch/fairseq}} and use the joint vocabulary as the pretrained model.
We compare with three groups of baselines:
\begin{itemize}
    \item \textbf{Auto-regressive model}: Transformers \cite{vaswani2017attention}. Which generates sentences in an auto-regressive manner. We follow the setting and the results introduced by \cite{gao2022difformer}, beam search with a beam size of 5 is used during generation.
    \item \textbf{Continuous Diffusion}: SeqDiffuSeq \cite{yuan2022seqdiffuseq}, DiffuSeq \cite{gong2022diffuseq}, Difformer \cite{gao2022difformer}. Which generates the sentences from a continuous latent space. We evaluate Difformer in the origin dataset without knowledge distillation on WMT14 dataset.
    \item \textbf{Discrete Diffusion}: CMLM \cite{ghazvininejad2019mask}, RDM \cite{zheng2023reparameterized}. Which generates sentences by denoising for each token gradually. We implement the RDM based on the same data setting and batch setting as our models.
\end{itemize}

\begin{table*}[ht]
\centering
\begin{tabular}{cccc}
\toprule
\textbf{Model} & \textbf{Type} & \textbf{IWTLS14 (De-En)} & \textbf{WMT14 (En-De)}  \\
\midrule
Transformer & Seq2seq & 33.91 & \textbf{27.37} \\ \midrule
Diffuseq (b=10) & Continuous Diffusion & 28.78  & 15.37   \\
Seqdiffuseq (b=10) & Continuous Diffusion & 30.03 & 17.14 \\
Difformer (b=20) & Continuous Diffusion & 34.13  & 21.42 \\ \midrule
RDM & Discrete Diffusion & \textbf{34.49} & 22.30 \\
CMLM & Discrete Diffusion & 31.76 & 20.03 \\ \midrule
XLDM (ours) & Discrete Diffusion & 23.78 & 20.30 \\
\bottomrule
\end{tabular}
\caption{Model Performance.}
\label{table:performance}
\end{table*}

\subsection{Experimental Settings}
\paragraph{Model Framework} We constructed our XDLM on an encoder-decoder architecture, with both the encoder and decoder comprising six Transformer layers. We set the hidden size of the model to 512 with eight attention heads.
\paragraph{Pretraining Stage Setting} During the pretraining stage, we formulate the pretraining task on the large-scale corpus mentioned above. We initialize pretraining with a weight decay rate of 0.0005 and a dropout rate of 0.2. We set the maximum number of tokens in each batch to 4k and provided 30k warm-up steps. Besides, the max length of both the source and target language is 256, aiming to make the input sentence a proper length.
\paragraph{Fine-tuning Stage Setting}
We apply fine-tuning on corresponding datasets, leveraging the robust foundation established from the pretraining datasets. The parameter setting for the fine-tuning process is primarily based on the pretraining stage, but with a smaller learning rate of 5e-5.

\subsection{Main result}
To ascertain the efficacy of pretraining on XDLM, we fine-tuned the model across several machine translation tasks, with comparative results in Table \ref{table:performance}. Our model demonstrates superior performance relative to certain continuous diffusion models, with the exception of Difformer. However, it exhibits comparable effectiveness to some discrete diffusion models, such as CMLM on the WMT14 dataset.

During the evaluation phase, we assess the BLEU score at both word and BPE (Byte Pair Encoding) levels, each requiring different tokenization scales. A comparison of the two tokenization methods is depicted in Table \ref{table:bleu}. Our findings indicate that our model performs more effectively when evaluated at the BPE level tokenization across two datasets, registering an improvement of approximately 5\%. This enhancement can be attributed to the fact that different tokenization levels help mitigate the complexity of the problems.

\begin{table}[ht]
\centering
\begin{tabular*}{0.35\textwidth}{lcc}
\toprule
 & \textbf{IWSLT14} & \textbf{WMT14} \\
\midrule
word level & 18.38 & 15.20 \\
BPE level & 23.78 & 20.30 \\
\bottomrule
\end{tabular*}
\caption{The BLUE results under different tokenization levels.}
\label{table:bleu}
\end{table}

%% file: ablation.tex
\section{Ablation Study}
\label{sec:ablation}
\begin{figure}[h]
\centering
\includegraphics[width=0.5\textwidth]{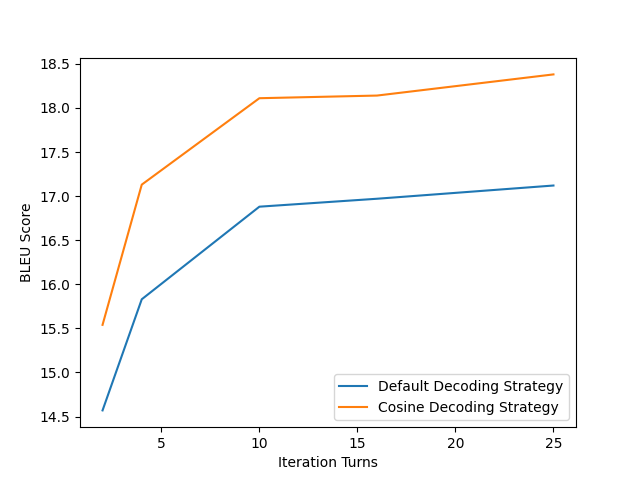}
\caption{
The effect of generation iterations on the BLEU score. }
\label{fig:figure 2}
\end{figure}

\textbf{Study on the Denoising Capacity at Intermediate Steps}
We also focus on the denoising capacity in each diffusion step. In the reverse process, XDLM apply $T$ diffusion steps to the Gaussian noise $Z_T$, generating correspond output $y$ after all intermediate steps. Figure \ref{fig:figure 2} shows the change of BLEU scores with the increasing of reverse steps. We can find for different decoding settings, our model can reach a stable result after 10 iterations, which shows out the effectiveness of our model.  

\textbf{Discussion}
In this section, we concentrate on the factors that contribute to the comparatively lower performance of our model relative to other models. 
One may have noticed that our method is not able to perform against the original RDM method, we discuss a few reasons in this section. Firstly, prior research such as RDM leverages a substantial batch size coupled with an extensive number of training iterations, a strategy that has been shown to enhance performance. Due to our machine limitations, we failed to conduct the experiments with the same level. Secondly, in terms of our pretraining configuration, we employ a pretraining dataset with an expanded vocabulary size to construct the Byte Pair Encoding (BPE) codes. This approach, while comprehensive, inadvertently increases the complexity of the problem and introduces out-of-vocabulary words that the model must interpret. Such challenges are not typically encountered in previous works. This discrepancy in methodology could potentially account for the performance differential observed.